\documentclass{clv3}

\usepackage{hyperref}
\usepackage{xcolor}

\usepackage{amsmath}
\usepackage{ulem}
\usepackage{commath}

\usepackage[russian,english]{babel}

\usepackage{graphicx}
\usepackage{float}
\usepackage{latexsym}
\usepackage{algorithm}
\usepackage{algorithmicx}
\usepackage{latexsym}
\usepackage{algpseudocode}
\usepackage{multicol}
\usepackage{amsmath}
\usepackage{enumitem}
\usepackage{multirow} 
\usepackage{tipa}

\usepackage{subcaption}

\usepackage{CJKutf8}
\usepackage{times}
\usepackage{latexsym}

\definecolor{darkblue}{rgb}{0, 0, 0.5}
\hypersetup{colorlinks=true,citecolor=darkblue, linkcolor=darkblue, urlcolor=darkblue}

\begin{document}
\issue{1}{1}{2016}

\dochead{Patterns of Persistence and Diffusibility across the World's Languages}

\runningtitle{Patterns of Persistence and Diffusibility}

\runningauthor{Chen and Bjerva}



\author{Yiyi Chen\thanks{E-mail: yiyic@cs.aau.dk.}}
\affil{Aalborg University, Copenhagen, Denmark}


\author{Johannes Bjerva}
\affil{Aalborg University, Copenhagen, Denmark}

\maketitle

\begin{abstract}
Language similarities can be caused by genetic relatedness, areal contact, universality, or chance.
Colexification, i.e.~a type of similarity where a single lexical form is used to convey multiple meanings, is underexplored. 
In our work, we shed light on the linguistic causes of cross-lingual similarity in colexification and phonology, by exploring genealogical stability (\textit{persistence}) and contact-induced change (\textit{diffusibility}). 
We construct large-scale graphs incorporating semantic, genealogical, phonological and geographical data for 1,966 languages.
We then show the potential of this resource, by investigating several established hypotheses from previous work in linguistics, while proposing new ones. 
Our results strongly support a previously established hypothesis in the linguistic literature, while offering contradicting evidence to another.
Our large scale resource opens for further research across disciplines, e.g.~in multilingual NLP and comparative linguistics.
\end{abstract}

\section{Introduction}

Comparative linguistics explores the intricate and diverse patterns that influence the current structure (synchrony) and historical development (diachrony) of languages. By examining these patterns and uncovering their causes and origins, we gain a deeper understanding of the similarities and differences between languages. Language similarities have been extensively employed in various field, such as multilingual natural language processing (NLP). In NLP, a commonly employed strategy is transfer learning, which involves leveraging the knowledge from high-resource languages to aid low-resource languages. Within this context, many studies have investigated the role of typology in NLP, e.g., via the similarity in syntactic typology across languages~\citep{malaviya-etal-2017-learning,bjerva2018phonology, bjerva2018tracking,
bjerva-augenstein-2021-typological,cotterell-etal-2019-complexity,bjerva-etal-2019-probabilistic,bjerva-etal-2019-uncovering, bjerva-etal-2020-sigtyp,
stanczak-etal-2022-neurons,ostling-kerfali-2023, 
fekete-bjerva-2023-gradual,bjerva2023role}. 
Recently work has also investigated semantic typology as a valuable resource in this area of study~\citep{chen-etal-2023-colex2lang,liu2023crosslingual}.

Within linguistics, recent work has investigated what factors govern typological patterns across languages. 
Colexifications represent one research direction, where the focus is on uncovering and exploring patterns across languages. 
There is often an underlying assumption that any shared colexification patterns are a result more from areal contact, rather than genealogical inheritance, with a general lack of empirical evidence ~\citep{haugen1950analysis,ross2001contact,ross2007calquing,matras2007investigating,haspelmath2009lexical}.
As an example, \citet{GastKoptjevskajaTamm+2022+403+438} investigate how colexification patterns and phonological forms of lexical items are genealogically inherited or acquired through language contact in Europe.
The main limitations in such work, are found in the scale of languages looked at, the binary modelling of the colexification patterns for language similarities, the reliance on an external database without context and low scalability of manual curation of the language contact networks.
Our work extends significantly upon this research direction, providing a comprehensive resource on which to build empirical evidence.

\paragraph{Contributions}

\begin{figure}[ht]
    \centering
    \includegraphics[width=0.5\textwidth]{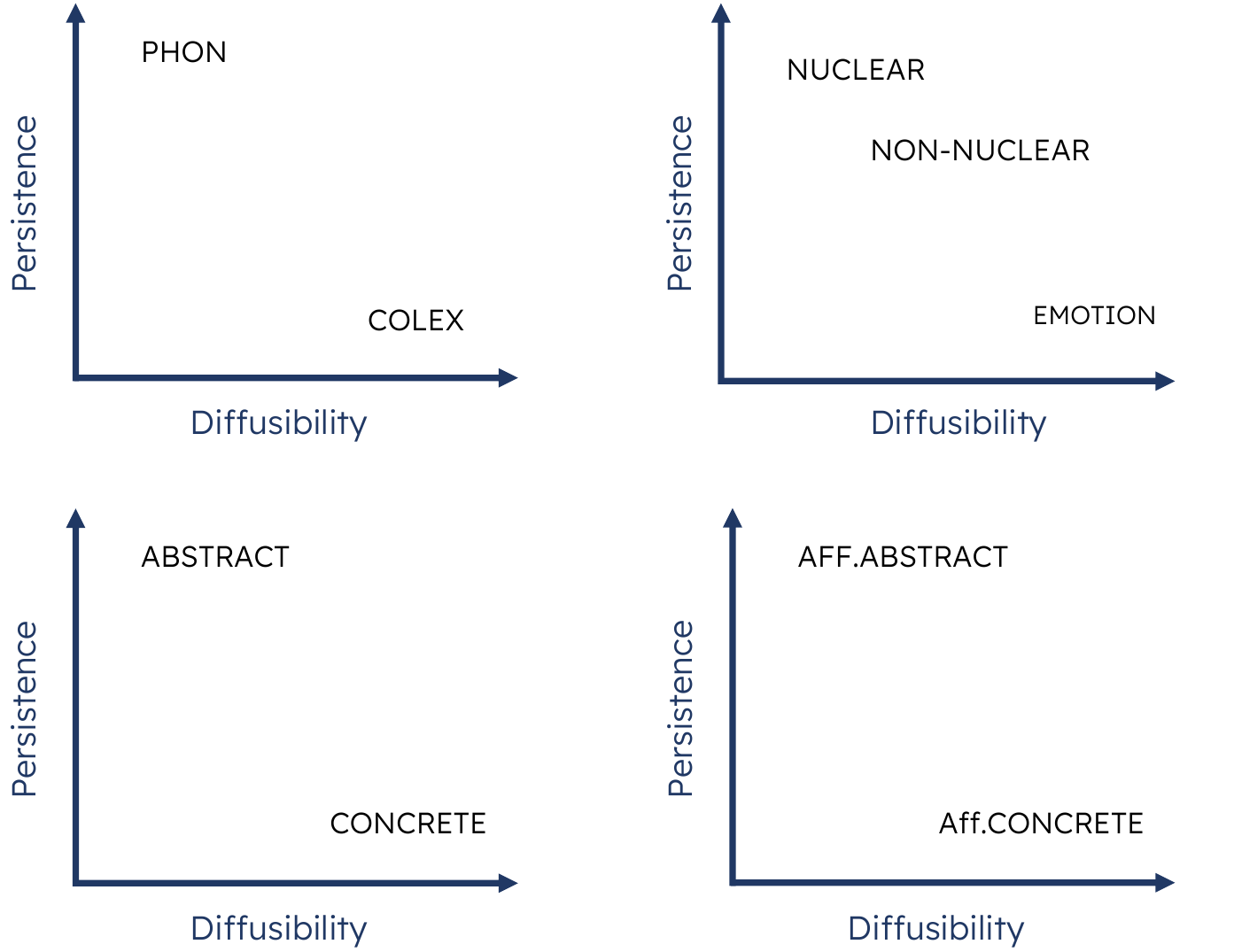}
    \caption{A visualization of the Hypotheses. Left Top: Hypothesis \textbf{H.1} of low persistence and high diffusibility of colexification patterns compared to phonological patterns. Right Top: Hypothesis \textbf{H2.a} of differential persistence and diffusibility in colexification patterns in nuclear, non-nuclear and emotion vocabularies. Left Bottom: Hypothesis \textbf{H2.b}  of high persistence and low diffusibility of abstract colexification patterns compared to concrete colexification patterns. Right Bottom: Hypothesis \textbf{H2.c}  of high persistence and low diffusibility of affectively loaded abstract colexification patterns compared to affectively loaded concrete colexification patterns.}
    \label{fig:hypos}
\end{figure}

We build large-scale language graphs incorporating semantic, phonological, and genealogical data, opening for a range of research questions in linguistics, with use cases more broadly in cross-lingual NLP\footnote{GitHub: ~\url{https://anonymous.4open.science/r/PersistenceAndDiffusibility-018F/}, OSF:~\url{https://osf.io/5sh62/?view_only=5d07119803c24743940a08777884cc33}}.
We test the following hypotheses, visualized in Fig.~\ref{fig:hypos}:

\begin{itemize}
    \item [\textbf{H1}.] \textbf{ Hypothesis of (a) Low Persistence and (b) High Diffusibility of Colexification Patterns}
    Degrees of similarity between languages in terms of colexification patterns involving genealogical stable concepts (\textsc{nuclear}) (a) reflect genealogical relatedness to a lesser extent than and (b) reflect language contact to a greater extent than degrees of similarity in terms of the phonological form of \textsc{nuclear} concepts do. ~\citep{GastKoptjevskajaTamm+2022+403+438}

    \item [\textbf{H2}.] Hypothsis of Differential Diffusibility and Persistence of Colexification Patterns of 
    \begin{itemize}
        \item [\textbf{H2.a}.] \textbf{Differential Genealogical Stability}
        Colexifications involving more genealogical stable concepts are more persistent and less diffusible than colexificaitons involving less genealogical stable concepts.~\citep{GastKoptjevskajaTamm+2022+403+438} 

        \item [\textbf{H2.b}.] \textbf{Abstractness and Concreteness}
        Colexifications involving abstract concepts are more persistent and less diffusible than colexificaitons involving concrete concepts. 
        
        \item [\textbf{H2.c}.] \textbf{Affectively loaded Abstractness and Affectively loaded Concreteness}
        Colexifications involving affectively loaded abstract concepts are more persistent and less diffusible than colexificaitons involving affectively loaded concrete concepts.

    \end{itemize}
\end{itemize}


By testing hypotheses \textbf{H1} and \textbf{H2.a} from previous work in linguistics, we show the potential of this resource -- interestingly, we verify \textbf{H1.a}, and contradict \textbf{H1.b} and \textbf{H2.a}. Furthermore, we showcase a suite of new hypotheses, \textbf{H2.b} and \textbf{H2.c}, on this new resource. Finally, we discuss the potential applications of language graph and conclude.

\section{Related Work}
\subsection{Colexification}

The concept of cross-linguistic colexifications was initially formulated by~\citet{franccois2008semantic} as a means of creating semantic maps. 
These maps visually depict the relationships between recurring expressions of meaning in a language~\citep{haspelmath2003geometry},
as language-specific colexification patterns indicate semantic proximity or connections between the meanings being colexified~\citep{hartmann2014identifying}. 
Exploring colexification patterns across languages provides valuable insights in various fields. 
For instance, it helps uncover cognitive principles~\citep{XU2020104280,BROCHHAGEN2022105179},
diachronic semantic shifts in individual languages~\citep{
karjus2021conceptual,François+2022+89+123}, and 
research in areal linguistics~\citep{koptjevskaja-tamm_liljegren_2017,SchapperKoptjevskajaTamm+2022+199+209,heine_kuteva2003}.

In the domain of emotions,~\citet{jackson-2019} conduct research on cross-lingual colexifications and discovered that distinct languages associate emotional concepts differently. 
For example, Persian speakers closely link the concepts of \textsc{grief} and \textsc{regret} whereas Dargwa speakers connect it with \textsc{anxiety}. 
This cultural variation and universal structure in emotion semantics is applied in NLP~\citep{sun-etal-2021-cross}.
For our study, we incorporate the concepts from emotion semantics to construct colexification patterns to test differential persistence and diffusibility.

\citet{di2021colexification} use colexificaiton patterns to test whether words connected through colexification patterns convey similar affective meanings. 
Inspired by this,~\citet{chen2023colexifications} create a large-scale concept graph incorporating concreteness, affectiveness and phonetic data leveraging colexification patterns.
\citet{liu2023conceptualizer} create a bipartite directed alignment graph across languages using colexification patterns. Built on which, \citet{liu2023crosslingual} create a multilingual graph ColexNet+. 

\subsection{Areal Linguistics}

In areal linguistics, the concept of horizontal transfer refers to language features affecting each other through contact across longer periods of time, e.g.~via borrowing lexical items.
Such lexical borrowings are relatively simple to analyze, and have been thoroughly explored in the literature.
Within the scope of language contact, shared features within geographically defined areas are also explored.
For instance, macroareas for the world's languages have been defined \citep{dryer1989large, dryer1992greenbergian} with the purpose of investigating, e.g., universals, while complementary work also points out how contact influences can also be found across boundaries of such areas \citep{miestamo2016sampling}.
Other types of area definitions also exist, e.g.~in AUTOTYP \citep{nichols2009autotyp}, while others propose purely radius-based measures \citep{jaeger2011mixed,cysouw2013chapter,bjerva-etal-2020-sigtyp}.

Within the context of areal linguistics, \citet{GastKoptjevskajaTamm+2022+403+438} investigate the degree to which colexification patterns are susceptible to diffusion, using the CLICS\textsuperscript{3} database to identify areal patterns in colexifications in Europe. 
In contrast, we use frequencies of colexification patterns from large-scale parallel data, indicating contextual information beyond the binary coding of patterns in~\citet{GastKoptjevskajaTamm+2022+403+438}.


\subsection{Resources}

Glottolog/Langdoc~\citep{hammarstrom2011langdoc} is a bibliographic database that catalogs languages worldwide, including lesser-known languages, assigning unique and stable identifiers to each language, family, and dialect. The World Atlas of Language Structures (WALS)~\citep{wals} offers structural information of Languages. 
\citet{littell2017uriel} constructs URIEL, which includes matrices representing language distances across various categories (e.g., geography, genealogy, syntax, phonology) for over 7,000 languages. 
Additionally, AUTOTYP\footnote{\url{https://github.com/autotyp/autotyp-data}}~\citep{bickel_balthasar_2023_7976754} supplies data for areas, whose definitions are free of linguistic information in order to avoid circularity in areal linguistics area~\citep{bickel2006oceania}. 
For our study, we extract and process genealogical, geographical, and syntactic data from these databases.

ColexNet+~\citep{liu2023crosslingual} is created using colexification patterns from Bible translations from the PBC corpus~\citep{mayer-cysouw-2014-creating} across 1,335 languages. 
In this study, we extract the frequency of colexifications in Bible data from ColexNet+ to model language similarities, to further examine the hypothesis of persistence and diffusibility of colexification patterns. The underlying assumption is that the widespread occurrence of colexification patterns across different languages signifies semantic similarities between those languages, which aligns with the frequentist perspective in linguistics~\citep{johnson2008quantitative}. With this approach, we can also avoid the shortcomings using an external translation-based database such as CLICS\textsuperscript{3}\citep{clics}, such as lacking of context and binary modeling on the colexification patterns.


Norms and ratings of words are essential components in psychology, linguistics, and increasingly used in NLP. 
Norms capture the typical usage frequency and context of words in a specific language, while ratings reflect subjective evaluations of individuals on dimensions like concreteness, valence, arousal, and imageability~\citep{brysbaert2014concreteness,warriner2013norms}. 
\citet{di2021colexification} show that colexification patterns capture similar affective meanings, indicating that concepts with closer affective associations are more likely to colexify.
Additionally, studies demonstrate a tendency for abstract words to co-occur with other abstract words and vice versa for concrete words~\citep{frassinelli-etal-2017-contextual, naumann-etal-2018-quantitative}
In our study, we extract concepts representing different degrees of concreteness and affectiveness to examine the differential persistence and diffusion, similar to the mapping process of \citet{chen2023colexifications}.

\subsection{Patterns of Persistence and Diffusability}

Similarities between languages can manifest across levels of linguistic structure, and are governed and affected by various factors.
For instance, if two languages exhibit phonological similarities in their core concepts, such as the English word \textit{water} and the Norwegian word \textit{vann}, both meaning~\textsc{water}, it may suggest a genealogical connection between the languages. 
While \begin{CJK}{UTF8}{min}\textit{コンピューター}\end{CJK} (Konpyūtā) in Japanese and \textit{der Computer} in German both stem from the English word~\textit{computer}, which is an example of language contact.

Previous research has relied on phonological similarities to identify the genealogical relatedness~\citep{
darquennes2006thomason, jager2018global}. This approach is based on the underlying assumption that certain concepts are universally entrenched in the languages and their corresponding words are resistant to change throughout language evolution, despite potential variations in their historical stability~\citep{holman2008explorations,
tadmor2010borrowability}. In such sense, core concepts from Swadesh list~\citep{swadesh1950salish} is presumed to exhibit substantial \textit{persistence}~\citep{swadesh1955towards} and limited susceptibility to \textit{borrowability}~\citep{carling2019causality}

Generalizations regarding borrowability and \textit{diffusibility} have primarily been formulated in relation to different levels of language system~\citep{matisoff2001genetic,wichmann2009assessing}. 
It is widely recognized that the lexicon is particularly prone to change induced by language contact~\citep{
kuteva2017contact}. 
However, the potential for ordering and transferring phonological patterns has not been extensively explored~\citep{Koptjevskaja-tamm2011}. 
Moreover, language contact as cause for lexico-semantic patterns is taken for granted in linguistic research, with few exceptions~\citep{smith1994mesoamerican,brown2011role,hayward1991propos,hayward2000there}.

We conduct a large-scale study on the persistence and diffusibility of colexification and phonological patterns.

\section{Constructing a Rich Language Graph}\label{language_graph}

\begin{figure}[htb]
    \centering
    \includegraphics[width=0.4\textwidth]{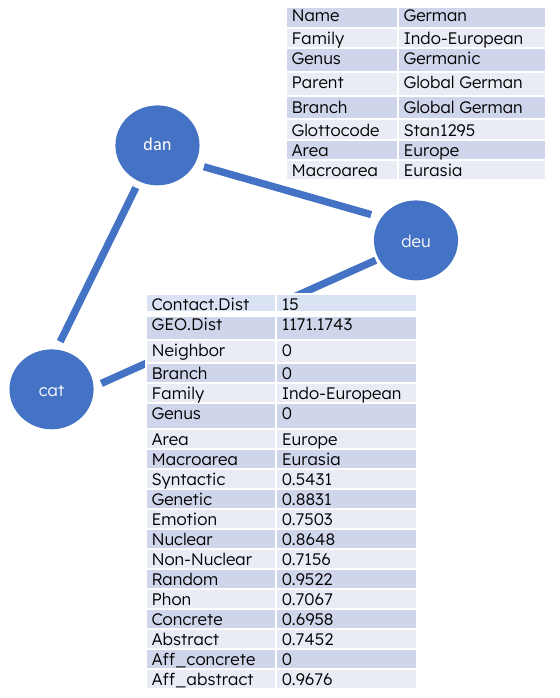}
    \caption{Language Graph. Each node represent a language with a table of attributes for language name, family, genus, parent, branch, glottocode, are and macroareas. Each edge represent distances for a pair of languages, in the aspect of geographical, language contact, areal categories, and other distances calculated from colexification patterns and phonological forms.}
    \label{fig:example}
\end{figure}

To examine persistence and diffusibility, we collect data from various sources to gauge language (dis)similarities across semantic and phonological spectra. We also consider language contact intensity and genealogical relatedness to construct a comprehensive language graph on a large scale. The specific steps involved in data curation and processing are outlined below. 
As a baseline for comparison with semantic and phonological distances, we include the \textsc{syntactic} distances from~\citet{littell2017uriel}. 
We use distance measures to quantify the relationships and degrees of difference between languages, in line with previous linguistic research. 
Fig.~\ref{fig:example} shows an example of the resulting language graph, where each node represents a language with its attributes, and the table associated with the edges represents the extracted and processed attributes used to capture language distances.

\begin{table*}[htb]
    \centering
  \resizebox{\textwidth}{!}{   
  \begin{tabular}{ccccc}
    \\\hline
              Language      & (\textsc{language}, \textsc{tongue}) &  (\textsc{eye}, \textsc{look}) &  (\textsc{tree}, \textsc{wood}) &  (\textsc{knee}, \textsc{kneel}) \\\hline 
Russian           &        \foreignlanguage{russian}{(язык, языках)}  163  &  \foreignlanguage{russian}{(гля, смотр, уви)} 264     &  \foreignlanguage{russian}{(дерев, инжир, пальм)} 228     &  \foreignlanguage{russian}{(колен)} 42 \\
Polish             &   (język, język) 169    &   0     &  (zawies, palm, drzew) 251     &    0 \\
Danish             & (tungemål, tunge, tungetale, tunger) 162  &     0    &   (træ, træe)  244       &  0 \\
German & (\"ubersetzen, schreibweise, zungen, zunge) 152 & 0 & 0&0 \\ 
Dutch & (taal, talen, tong) 158 & 0 & 0& 0\\
\hline

    \end{tabular}}
    \caption{Four colexifications in Russian, Polish, Danish, and Rundi, with their realization word forms and frequencies in the regarding languages. }
    \label{tab:colex_example}
\end{table*}

\paragraph{Concepts}
To test the differential persistence and diffusibility of colexification patterns across languages, we curate concepts of different degrees of genealogical stability, concerteness and affectiveness, for the construction of three colexification matrices. See details of the curated concepts in Appendix~\ref{sec:lexicon_appendix}.

First, for testing \textbf{H1} and \textbf{H2.a}, we first curate the \textsc{nuclear} concepts, which are the 40 most genealogically stable items of core vocabulary. For constructing colexification matrices for \textbf{H2.a}, we further curate concepts in \textsc{non-nuclear} (the rest of swadesh list), \textsc{emtion}~\citep{jackson-2019} and \textsc{random} (randomly sampled concepts disjunctive with the aforementioned concept lists), representing from the more genealogical stable to the lesser.


In~\citet{brysbaert2014concreteness}, 40,000 English word lemmas were rated on a scale of 1 to 5 for concreteness. Second, two concept lists are curated for testing \textbf{H2.b}, with concepts in the \textsc{abstract} set having ratings below 3, and the concepts in the \textsc{concrete} set having ratings above 4.


In the study by~\citet{warriner2013norms}, around 14,000 English lemmas were assigned valence, arousal, and dominance ratings on a scale of 1 (happy [excited; controlled]) to 9 (unhappy [calm; in control]). We define concepts with all three ratings below 4 or above 6 are as having high-level affectiveness, in consistency with~\citet{warriner2013norms}. To further examine the impact of affectiveness on differential persistence and diffusibility (\textbf{H2.c}), additional filtering was applied to the \textsc{abstract} and \textsc{concrete} lists. This filtering focused on the highly affectively loaded concepts, resulting in refined concepts referred to as \textsc{aff.abstract} and \textsc{aff.concrete}, respectively.

\paragraph{Colexification Matrices} 
In each language, a colexification pattern can be manifested by different word forms. 
For example, \textit{ähnlich} and \textit{vergleichen} in German are word forms that represent the colexification pattern (\textsc{compare},~\textsc{like}). To model the semantic distances among languages using colexifications, we construct colexification matrices based on the curated concept lists and the frequencies of the colexification patterns.

Similar to~\citep{GastKoptjevskajaTamm+2022+403+438}, 
we select the colexification patterns $(C_1, C_2)$ based on a set of concepts $\mathbf{C}$, where at least one concept, $C_1$ or $C_2$, is consisted in the colexification pattern in a language $L$. For example, as shown in Table~\ref{tab:colex_example}, the extracted four colexification patterns, in which the concepts \textsc{tongue}, \textsc{eye}, \textsc{tree}, and \textsc{knee} are included in the \textsc{nuclear} concept list, while the other concepts are not (cf. Appendix~\ref{sec:lexicon_appendix}). 

Contrary to~\citet{GastKoptjevskajaTamm+2022+403+438}, instead of modeling colexification patterns as binary attributes, i.e., occurring in a language or not, we take a \textit{distributional semantic} view on colexification patterns, for which we use ColexNet+ trained on Bible data~\citep{liu2023crosslingual}. Based on the aggregated frequencies of the realized word forms of a colexification pattern, we construct a vector $vec_L$ for each language $L$. For example, the word forms \textit{træ} and \textit{træe}, which are the realizations of the colexification pattern (\textsc{tree}, \textsc{wood}), occur 91 and 153 times in the Danish Bible corpus, respectively. As a result, we record the frequency of the colexification pattern (\textsc{tree}, \textsc{wood}) in Danish as 244, as illustrated in~Table~\ref{tab:colex_example}. In notation, each data point is formulated as $(C_1, C_2, L, F)$, where $C_1$ and $C_2$ are concepts, $L$ is a language, if the Bible data contains forms for $C_1$ and $C_2$ in $L$, and $F$ is a variable indicating how frequent the $C_1$ and $C_2$ are colexified in $L$. We also include the negative samples, where $F$ equals to $0$ if the colexification pattern does not occur in $L$. To model the language dissimilarities, each language $L$ is represented by a vector $v_L$ consisting of frequencies of all the occurring colexification patterns in $L$. Subsequentially, a colexification matrix consisting of a set of $n$ languages ${L_1, L_2, \dots, L_n}$ based on a set of concepts $\mathbf{C}$ is defined as $\mathbf{M}_{\mathbf{C}} = [v_{L_1}, v_{L_2}, \dots, v_{L_n}]$, where there are $m$ colexification patterns based on $\mathbf{C}$ and $\mathbf{M}_{\mathbf{C}}\in\mathbf{N}^{n,m}$. For example, considering only four colexification patterns in Table~\ref{tab:colex_example}, $v_{Russian} = [163, 264, 228, 42]$ and $v_{Polish}= [169, 0, 251, 0]$. Based on Table~\ref{tab:colex_example}, a colexification matrix can be constructed as $\mathbf{M}_{5,4}$=
$[\begin{smallmatrix}
    163 & 264 & 228 & 42 \\
    169 & 0  & 251 & 0  \\
        & \cdots & &     \\
    158 &0  & 0 &  0  \\
\end{smallmatrix}]$, consisting of 4 colexification patterns across 5 languages.
It is obvious in this example, German is more similar to Dutch than to Russian in terms of the overlapping colexification patterns. Taking the frequency into consideration, we compare languages sharing the same colexification patterns with various frequencies. To ensure the fairness for experimentation, the sets of colexification patterns include mutually exclusive concept lists, resulting in evenly distributed and disjointed colexification matrices (cf. Appendix~\ref{sec:lexicon_appendix}).

\paragraph{Semantic Distances}
By utilizing colexification matrices obtained from different concept sets, the semantic distances can be determined using the formula $1-\frac{<v_{L_i}, v_{L_j}>}{||v_{L_i}||\cdot||v_{L_j}||}$, ranging from $0$ (least distant) to $1$ (most distant). Here, $v_{L_i}$ and $v_{L_j}$ denote the language vectors corresponding to languages $(L_i, L_j)$. 
Fig.~\ref{fig:example} shows the calculated semantic distance using the \textsc{nuclear}-matrix labeled as \textsc{nuclear} in the language network. In the same vein, \textsc{non-nuclear}, \textsc{emotion}, \textsc{random}, \textsc{concrete}, \textsc{abstract}, \textsc{aff.concrete} and \textsc{aff.abstract} are calculated using the corresponding matrices. Overall, the semantic distances measured by colexification patterns are extracted for 1320 languages and 870,540 language pairs.

\paragraph{Phonological Distances} Using phonetically transcribed word lists from the ASJP database~\citep{asjp2022}, phonological distances between languages are determined based on the overall dissimilarity of 40 \textsc{nuclear} concepts~\citep{jager2018global}. These distances, measured by sound classes, are extracted for 1,558 languages and 1,212,903 language pairs. The phonological distances ~\textsc{phon} are used to populate the language graph, ranging from $0$ (most similar) to $1$ (most distant).

\paragraph{Genealogical Relatedness}
The genealogical data for each language is extracted from Glottolog~\citep{harald_hammarstrom_2022_7398962}\footnote{We use the most recent version Glottolog 4.7:~\url{https://zenodo.org/record/7398962}.}, such as \textsc{family}, \textsc{parent}, \textsc{branch}, and \textsc{macroarea} while \textsc{genus} is extracted from WALS~\citep{wals}. 
We record the each relationship between each pair of languages as either its name (identical) or 0 (un-identical), see Fig.~\ref{fig:example}. 
We define two languages are lower-level related, if they have the identical language branch; mid-level related with identical genus; higher-level related with identical family; otherwise unrelated. 
Furthermore, \textsc{genetic} distances among languages are extracted for further analysis, which are calculated based on the Glottolog language tree~\citep{littell2017uriel}. 
In addition, ~\textsc{area} data is extracted from AUTOTYP.

\paragraph{Linguistically Motivated Contact Intensity} Coordinates of languages are extracted from Glottolog database, with which the geodesic distances (KM)\footnote{\url{https://geopy.readthedocs.io/en/stable/index.html?highlight=geodesic\#geopy.distance.geodesic}} among languages are calculated, recorded as \textsc{geo.dist}. Furthermore, language contact distance, i.e., \textsc{contact.dist}, is measured as the number of languages in between each language pair based on geographical distances~\citep{dryer2018order}. 
While~\citet{GastKoptjevskajaTamm+2022+403+438} uses \textsc{geo.dist} to define language neighbours distinguishing national and minor languages, their language network is limited to European countries and difficult to scale across the world without comprehensive meta data.
For better scalability and consistency, we define neighbouring languages when there are less than 10 contact languages in between, recorded as a binary value 0 (non-neighbour) or 1 (neighbour) as~\textsc{neighbour}, align with~\citet{dryer2018order} (cf. Appendix~\ref{sec:languge_contact_appendix} for examples of language contact graphs). For experimentation, both~\textsc{contact.dist} and~\textsc{geo.dist} are rescaled and normalized to the range of 0 to 1.


\begin{figure*}[t]
    \centering
    \includegraphics[width=0.8\textwidth]{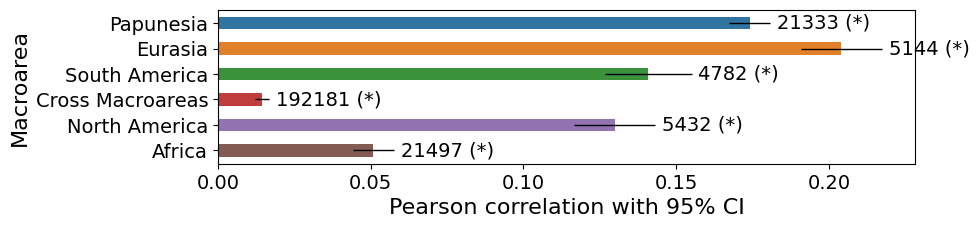}
    \caption{Pearson Correlation between Phonological Distances and Semantic Distances Per Macroarea. The numbers besides the bar indicates the number of language pairs (samples) per each macroarea. \textbf{Cross Macroareas} represent the language pairs are not in the same macroarea. All $p$-values are $<0.001$(*).}
    \label{fig:pearsonr_macroarea}
\end{figure*}

\section{Experiments, Analyses and Results}

Based on this rich language graph, we analyze the language dissimilarities measured by colexification and phonological patterns, and investigate established hypotheses \textbf{H.1} and \textbf{H.2.a} from linguistics \citep{GastKoptjevskajaTamm+2022+403+438}, building on areal linguistics, in addition to providing new hypotheses \textbf{H.2.b} and \textbf{H.2.c}.

\subsection{Phonological vs. Semantic Distances of Languages}

To examine the relationship between phonological and semantic dissimilarities among languages, Pearson correlations are computed between semantic distances (\textsc{nuclear}-matrix) and phonological distances (\textsc{phon}), both based on nuclear vocabulary. 
The correlations are analyzed within and across macroareas, where a dataset of 912 languages and 250,369 language pairs (samples) is used. 
A significant portion of language pairs (190,013) from different macroareas exhibit relatively low correlations (Fig.~\ref{fig:pearsonr_macroarea}).
However, across macroareas, there is a consistent increase in phonological distance as semantic distance increases, with all correlations being statistically significant. 
Notably, although {\color{blue}Papunesia} has a greater number of language pairs (21,333) compared to other macroareas, it also displays a stronger effect size ($r$) in terms of correlation.


\subsection{Language Distances of Phonology/Colexification Patterns vs. Language Contact Distances across Genealogical Relatedness}\label{subsec:phon_colex_geo}

\begin{figure*}[t]
    \centering
    \includegraphics[width=\textwidth]{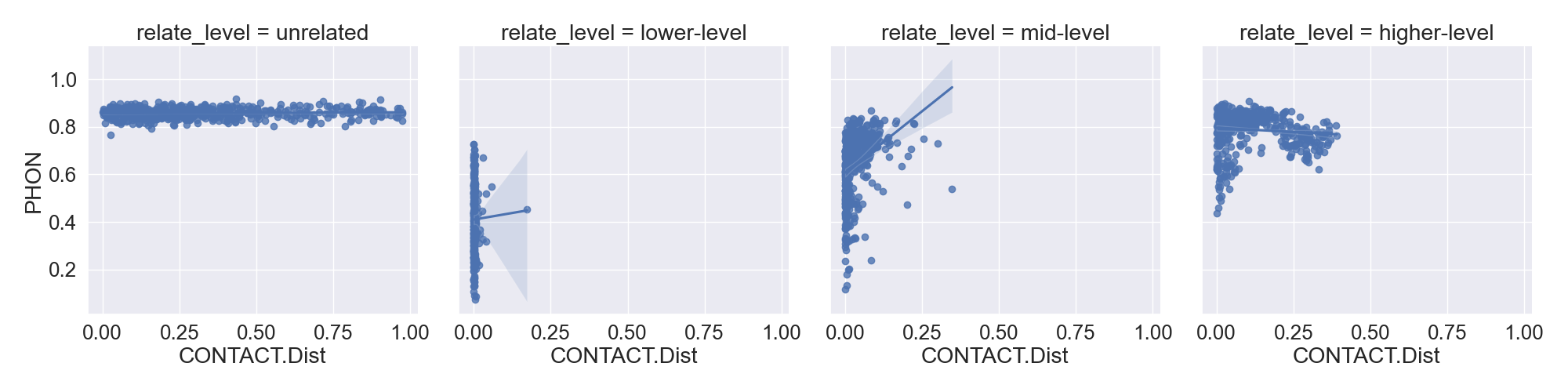}
    \includegraphics[width=\textwidth]{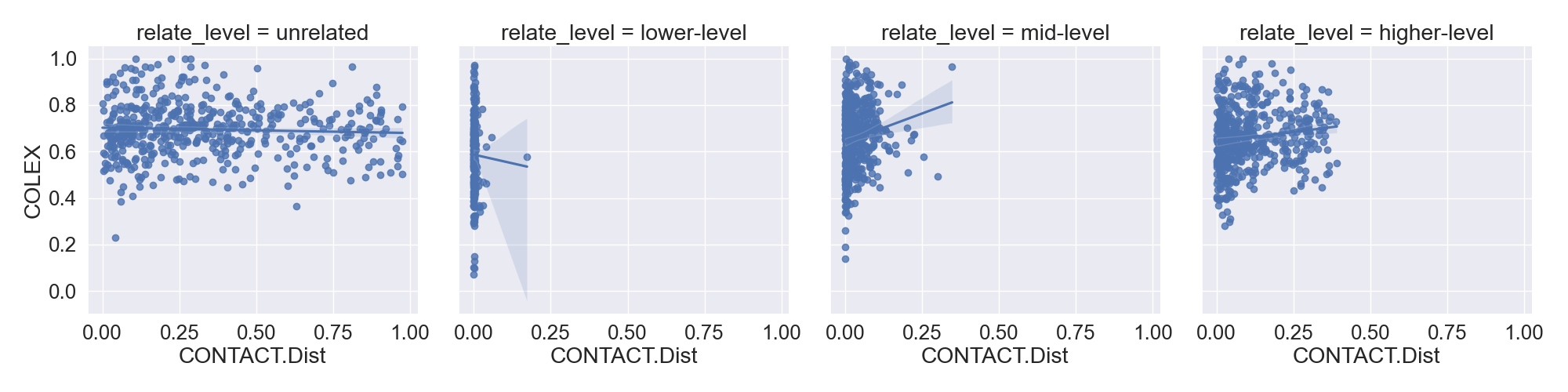}
    \caption{Phonological distances (top) and Semantic Distances (\textsc{nuclear}) (bottom) plotted against geographical distance, for four groups of genealogical relatedness (unrelated, lower-level related, mid-level related, and higher-level related). 
    }
    \label{fig:phon_colex_syn_lm}
\end{figure*}

Based on our data, we observe that the relationship between genealogical distance and contact distance as indicators of phonological distance is negative. 
In other words, when languages are closely (lower-level and mid-level) related, the correlation between contact distance and phonological distance is more pronounced compared to languages that are more distantly (higher-level) related or unrelated.



To investigate the relationship between language contact distances (\textsc{geo.dist}) and semantic/phonological distances in our data, we conduct a stratified sampling of 1,744 language pairs out of a total of 568,286 pairs with known relatedness, and employ linear regression. 
Fig.~\ref{fig:phon_colex_syn_lm} shows the phonological distance (top) and semantic distance (bottom) against contact distances
Each dot on the plots represents a language pair, divided into unrelated, lower-level related, mid-level related, and higher-level related language pairs, from left to right.

In terms of phonological distances, the slope of regression line is steepest for mid-level related languages, indicating a stronger correlation between contact distance and phonological distance. 
While for colexification patterns, they show a closer correlation with contact distance as well for mid-level languages, while a reverse correlation for lower-level contact distances, which contracts the observations made by~\citet{GastKoptjevskajaTamm+2022+403+438}. 
We note that their study is limited to a total of 45 languages, where they observe a notable correlation between colexification distances and contact distances for unrelated languages. 
Additionally, we observe notable disparities in variances. 
Phonological distances among higher-level related or unrelated languages consistently exhibit high values with minimal variance. On the other hand, phonological distances among closely related languages display greater variances, as do colexification distances across different levels of genealogical relatedness. 

It is notable that semantic distances decrease as language contact distances increase for language related at a lower level. To investigate further, the relationships between semantic distances and language contact distances for each branch are shown in Fig.~\ref{fig:colex_contact_lm_negative} and Fig.~\ref{fig:colex_contact_lm_positive} (cf. Appendix~\ref{sec:colex_branches}). Both negative and positive correlations are presented for different branches involving extremely low-resource languages, which can be inviting for further research.

\subsection{Persistence and Diffusibility of Phonology vs. Colexifications}\label{phon_colex}

\begin{figure}[t]
    \centering
    \includegraphics[width=0.5\textwidth]{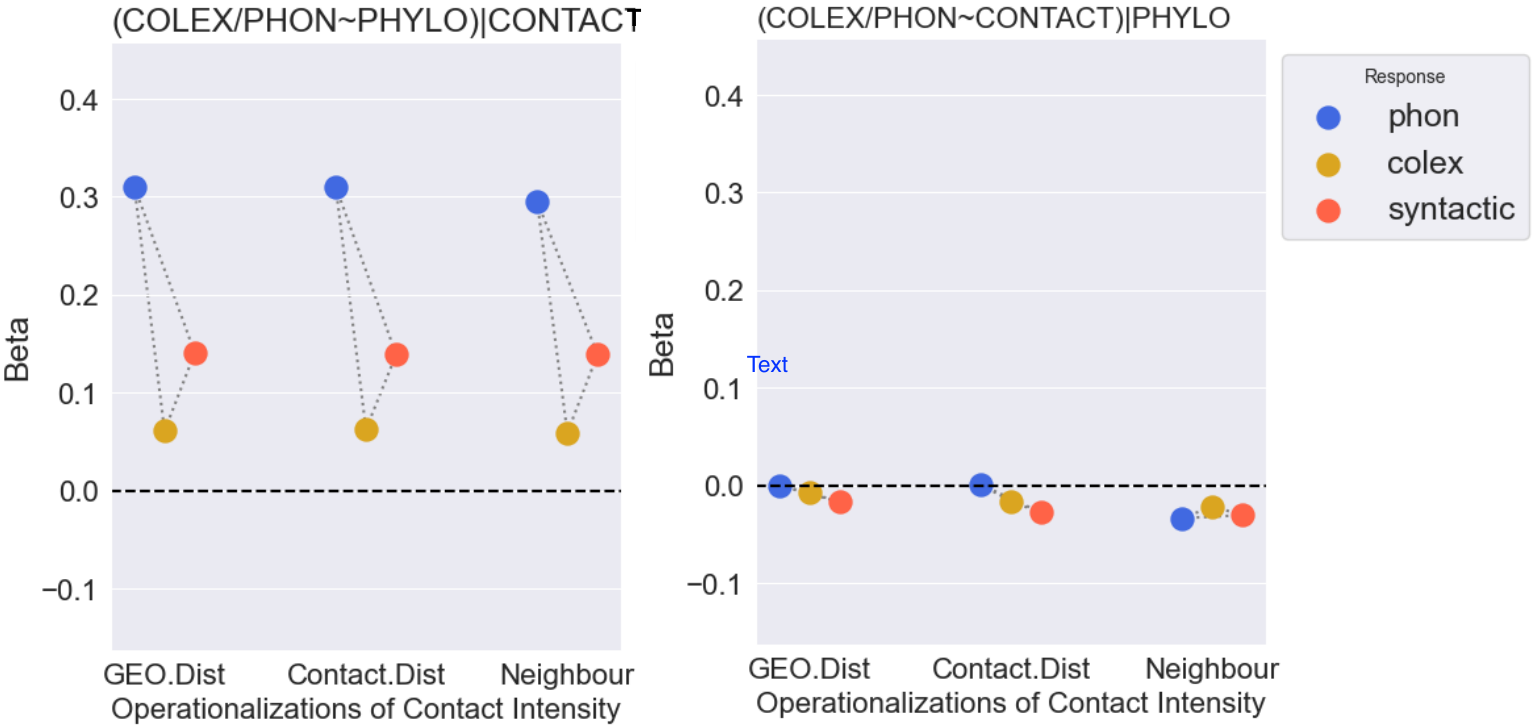}
    \caption{Beta Coefficients of Regression Analyses. The plot on the left shows the values for \textsc{phon} and \textsc{phylo} (blue), \textsc{colex} and \textsc{phylo} (yellow) and \textsc{syntactic} and \textsc{phylo} (orange) , with the three operalizations of contact intensity as control variables. The plot on the right shows the mean Beta coefficients for the three \textsc{contact} distances and the \textsc{phon} (blue), in comparison to the \textsc{colex} (yellow) and \textsc{syntactic} (orange) distances. The confidence interval at 95\% are within $(-0.05, 0.2)$ for Beta coefficients. $p<0.001$ for all.  }
    \label{fig:phon_colex_mixed}
\end{figure}

To examine the hypotheses \textbf{H.1} of persistence and diffusibility of colexification patterns and phonological forms, mixed effects regression modelling is implemented to test the correlations between variables of interest. We report the average Beta coefficients, indicating the magnitude of the effects and reflecting the strength of the correlations, for all language pairs in the sample. 

As depicted in Fig.~\ref{fig:phon_colex_mixed} (left), we compare the correlations between phylogenetic distance (\textsc{genetic}) and (i) colexification, (ii) phonological, and (iii) syntactic distances, while controlling for language contact intensity (\textsc{geo.dist}, \textsc{contact.dist}, and \textsc{neighbour}), grouped by the level of relatedness.

The dots in Fig.~\ref{fig:phon_colex_mixed} represent the mean values of Beta coefficients for 250,369 language pairs (912 languages), where we have data points of \textsc{nuclear}, \textsc{phon}, \textsc{syntactic}, and the level of relatedness. These values demonstrate the degree of persistence. The dashed lines connecting the dots represent the differences between paired sets of Beta coefficients.

The result confirms \textbf{H.1(a)} that phonological forms are more persistent than colexification patterns, as the Beta coefficients for \textsc{phon} and phylogenetic distance are significantly higher than those for both \textsc{colex} and \textsc{syntactic} with phylogenetic distance. In addition, \textsc{syntactic} exhibits a higer level of persistence compared to \textsc{colex}, although the differences are within 0.1 for all three control language contact intensity variables ($p<0.001$), corroborating the intuition that the syntactic features are more resistant to change than colexification patterns across languages.

Fig.~\ref{fig:phon_colex_mixed} (right) compares the correlations between the three language contact intensity variables and (i) colexification, (ii) phonological, and (iii) syntactic distances, while controlling phylogenetic distance (\textsc{genetic}). The mean Beta coefficients are near zero, compared to the left plot, and the difference is negligible, contradicting \textbf{H.1(b)}.
In contrast to~\citet{GastKoptjevskajaTamm+2022+403+438}, where only unrelated languages are considered for testing degrees of diffusibility and mid-high distant languages are tested for persistence, our study implement regression modelling on all levels of genealogical relatedness and all range of geographical distances.



The findings strongly support the hypothesis that colexification patterns exhibit low persistence, regardless of the control conditions for contact intensity. On the other hand, the hypothesis of higher diffusibility for colexification patterns is not well supported. 
Language distances derived from colexification patterns, much like those based on phonological forms, predominantly reflect the degree of genealogical inheritance rather than contact intensity (Fig.~\ref{fig:phon_colex_mixed}).

\subsection{Differential Persistence and Diffusibility of Colexification Patterns }
To test differential persistence and diffusibility of colexification patterns, we propose a suite of hypotheses, involving lexicons on different scales of genealogically stability, concreteness and affectiveness. We follow the same analysis as in Section~\ref{phon_colex} to test all the hypotheses in this section. For all the regression modelling, the languages are grouped by different levels of relatedness, and for all the reported Beta coefficients, the confidence intervals at 95\% are within $(-0.1,0.1)$ and $p<0.001$.


\begin{figure}[t]
    \centering
    \includegraphics[width=0.5\textwidth]{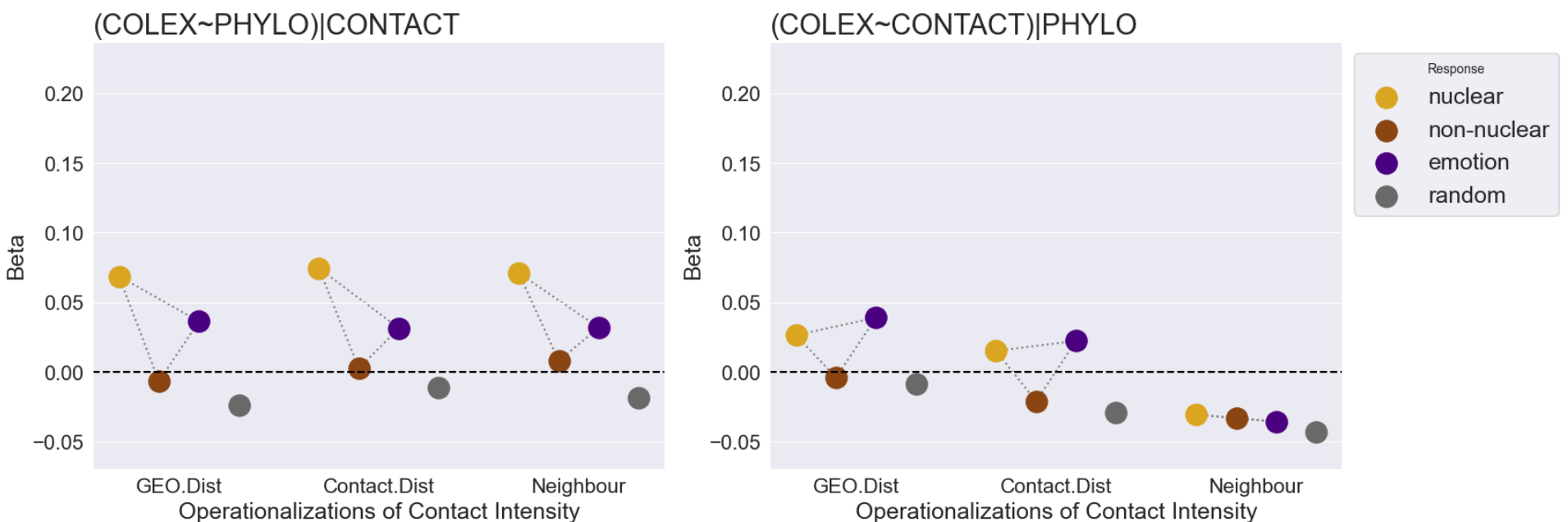}
    \caption{Beta Coefficients of Regression Analyses. The plot on the left shows the values for \textsc{colex.nuclear}, \textsc{colex.non-nuclear}, \textsc{colex.emotion}, and \textsc{colex.random}. The plot on the right shows the values for the three \textsc{colex}-matrices and the three \textsc{contact} distances.}
    \label{fig:colex1}
\end{figure}

\paragraph{Core vs. Emotion vs. Random Concepts }
Similar to previous research~\citep{GastKoptjevskajaTamm+2022+403+438}, we hypothesize $\textbf{H.2.a}$ that there is a differential persistence of colexification patterns involving core (\textsc{nuclear} and \textsc{non-nuclear}) concepts, \textsc{emotion} and \textsc{random} concepts.


As shown in Fig.~\ref{fig:colex1}, contrary to previous findings, the \textsc{colex.nuclear} are clearly more persistent and not less diffusible than the \textsc{colex.non-nuclear} concepts. Although the Beta coefficients are small, there is evidence supporting a gradual differential persistence from \textsc{colex.nuclear}, \textsc{colex.emotion}, \textsc{colex.non-nuclear} to \textsc{colex.random}, under all three control conditions for contact intensity. It is notable that, \textsc{colex.emotion} shows both more persistence and diffusibility than \textsc{colex.non-nuclear}. We further investigate cross-family language similarity using adjusted rand indices (ARIs) (cf. Appendix~\ref{pairwise_aris}).

\begin{figure}[tb]
    \centering
    \includegraphics[width=0.5\textwidth]{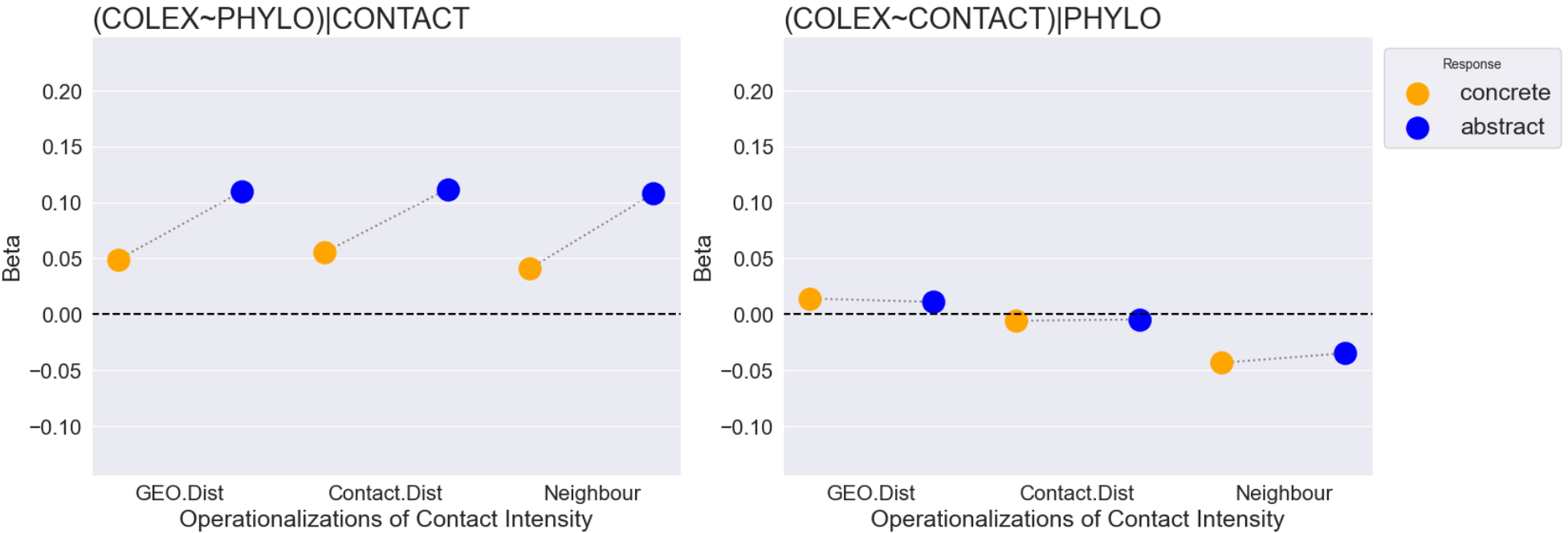}
    \caption{Beta Coefficients of Regression Analyses. The plot on the left shows the values for \textsc{colex.concrete} (orange), in comparison to \textsc{colex.abstract} (blue). The plot on the right shows the values for the two \textsc{colex}-matrices and the three \textsc{contact} distances.}
    \label{fig:colex2}
\end{figure}

\paragraph{Concreteness vs. Abstractness}
Previous research shows that concrete concepts are more easily learned and remembered than abstract concepts, and that language referring to concrete concepts is more easily processed~\citep{schwanenflugel2013abstract,guasch2021emotion}. Filtering the data with concreteness ratings, we observe that there are many more colexification patterns involving abstract concepts compared to concrete concepts with the same amount of concepts (as shown in Table~\ref{fig:dist_colex}), which aligns with the findings from large-scale empirical analyses of abstract and concrete concepts~\citep{hill2014quantitative}. 
Motivated by which, we hypothesize that the colexifications involving only concrete concepts are \textit{less persistent} and \textit{more diffusible}  than the colexifications involving only abstract concepts. 
However, Fig.~\ref{fig:colex2} shows certain support for the former, not the latter. 
The differences of Beta are rather weak for both correlations between colexifications and contact intensity, suggesting that concrete and abstract colexifications are equally likely to come into being through genealogical inheritance rather than language contact. 


Furthermore, affectively loaded abstract concepts are shown easier to acquire than affectively loaded concrete concepts~\citep{guasch2021emotion}. 
Interestingly, we also observe an inverse correlation between the phylogenetic relatedness and affectively loaded concrete colexifications (\textsc{colex.aff.concrete}), while there is a positive correlation between phylogenetic relatedness and affectively loaded abstract colexifications (\textsc{colex.aff.abstract}). The differential diffusibility between affectively loaded colexifications is more significant than without affectiveness.

\section{Conclusion and Future Work}

We construct large-scale language graphs that incorporates semantic, genealogical, phonological, and geographical data for overall 1966 languages and 1,931,595 language pairs. 
This resource allows us to explore several established hypotheses from previous linguistic research while proposing new ones.
Our findings strongly support one hypothesis established in the linguistic literature, we encounter contradictory results concerning another hypothesis. 
These results contribute to the ongoing discussion and provide valuable insights for further research in various disciplines, including multilingual natural language processing (NLP) and comparative linguistics. For future work, language similarities presented in the graphs will be directly used as guideline for cross-lingual transfer learning, and as features to be employed for parameter composition in adapters to improve downstream tasks in zero-shot learning~\citep{ansell-etal-2021-mad-g}.
In conclusion, We offer a comprehensive analysis of genealogical stability and contact-induced change in colexification and phonology across a wide range of languages. 
The large-scale language graphs we present serve as valuable resources for future interdisciplinary research, bridging the gap between multilingual NLP and comparative linguistics. 

\section*{Limitations}

This work generally relies on pre-existing linguistic resources, some of which have been gathered with an English or Indo-European bias.
Furthermore, the distributional colexifications we extract are based on bible data, which constitutes translated material.
As the domain of lexical forms is peculiar we do expect there to be a domain-specific effect.
However, we hypothesize that this effect is minor as it should primarily affect \textit{which} colexifications we find, and not their distribution as such. 
Consider the lexical items in a language V, a subset of its items $V_{bible}$ and its complement $V_{non-bible}$. 
While our specific domain only allows us to explore distributional colexifications in the context of $V_{bible}$, these are arguably the same lexical items that we will find in other languages. 
Hence, the view we present has a narrower scope due to this domain effect, but we expect, e.g., the cosine distance between two languages to be highly similar regardless of domain. 
Nonetheless, building these patterns directly on non-translated language resources would constitute an improvement, albeit infeasible given available textual resources.

\section*{Ethics Statement}

The ethical issues in this work are negligible, as we develop a linguistic resource and perform experiments using it. No new human data was gathered in this process.
There is a minor risk, that the Indo-European bias in some of these resources, and in Bible translations, can result in further propagating western biases.

\section{Acknowledgments}
This work is supported by the Carlsberg Foundation under a \textit{Semper Ardens: Accelerate} career grant held by JB, entitled ``Multilingual Modelling for Resource-Poor Languages'', grant code CF21- 0454. 
We furthermore thank Esther Ploeger for her assistance in calculating the contact distances among languages.

\starttwocolumn
\bibliography{compling_style}

\appendix


\appendixsection{Concepts}\label{sec:lexicon_appendix}

\paragraph{\textsc{nuclear}}
\textsc{blood}, \textsc{bone}, \textsc{breast}, \textsc{come}, \textsc{die}, \textsc{dog}, \textsc{drink}, \textsc{ear}, \textsc{eye}, \textsc{fire}, \textsc{fish}, \textsc{full}, \textsc{hand}, \textsc{hear}, \textsc{horn}, \textsc{knee}, \textsc{leaf}, \textsc{liver}, \textsc{louse}, \textsc{mountain}, \textsc{name}, \textsc{new}, \textsc{night}, \textsc{nose}, \textsc{one}, \textsc{path}, \textsc{person}, \textsc{see}, \textsc{skin}, \textsc{star}, \textsc{stone}, \textsc{sun}, \textsc{tongue}, \textsc{tooth}, \textsc{tree}, \textsc{two}, \textsc{water}, \textsc{i}, \textsc{you}, \textsc{we}

\paragraph{\textsc{non-nuclear}}

\textsc{all}, \textsc{ash}, \textsc{bark}, \textsc{belly}, \textsc{big}, \textsc{bird}, \textsc{bite}, \textsc{black}, \textsc{burn}, \textsc{claw}, \textsc{cloud}, \textsc{cold}, \textsc{dry}, \textsc{earth}, \textsc{eat}, \textsc{egg}, \textsc{feather}, \textsc{flesh}, \textsc{fly}, \textsc{foot}, \textsc{give}, \textsc{good}, \textsc{grease}, \textsc{green}, \textsc{hair}, \textsc{head}, \textsc{heart}, \textsc{hot}, \textsc{kill}, \textsc{know}, \textsc{lie}, \textsc{long}, \textsc{man}, \textsc{many}, \textsc{moon}, \textsc{mouth}, \textsc{neck}, \textsc{not}, \textsc{rain}, \textsc{red}, \textsc{root}, \textsc{round}, \textsc{sand}, \textsc{say}, \textsc{seed}, \textsc{sit}, \textsc{sleep}, \textsc{small}, \textsc{smoke}, \textsc{stand}, \textsc{swim}, \textsc{tail}, \textsc{that}, \textsc{this}, \textsc{walk}, \textsc{what}, \textsc{white}, \textsc{who}, \textsc{woman}, \textsc{yellow}

\paragraph{\textsc{emotion}}
\textsc{bad}, \textsc{want}, \textsc{love}, \textsc{hate}, \textsc{happy}, \textsc{grief}, \textsc{shame}, \textsc{fear}, \textsc{anger}, \textsc{envy}, \textsc{proud}, \textsc{regret}, \textsc{anxiety}, \textsc{pity}, \textsc{surprised}, \textsc{hope}, \textsc{like}, \textsc{sad}, \textsc{merry}, \textsc{joy}, \textsc{desire}, \textsc{gloomy}, \textsc{worry}

\paragraph{\textsc{random}}
\textsc{understand}, \textsc{glorying}, \textsc{horned}, \textsc{mahli}, \textsc{hadad}, \textsc{zippor}, \textsc{petition}, \textsc{wretched}, \textsc{sapphire}, \textsc{beam}, \textsc{gob}, \textsc{sheshai}, \textsc{aloud}, \textsc{bud}, \textsc{singer}, \textsc{slingstone}, \textsc{glutton}, \textsc{field}, \textsc{morrow}, \textsc{allegory}, \textsc{enter}, \textsc{city}, \textsc{arabia}, \textsc{harp}, \textsc{proverb}, \textsc{medeba}, \textsc{pelethites}, \textsc{brass}, \textsc{adoniram}, \textsc{flour}, \textsc{fleece}, \textsc{apple}, \textsc{suffer}, \textsc{stubborn}, \textsc{net}, \textsc{thunder}, \textsc{stomach}, \textsc{visitation}, \textsc{crying}, \textsc{face}, \textsc{handful}, \textsc{abishag}, \textsc{uz}, \textsc{hushai}, \textsc{chew}, \textsc{theirs}, \textsc{ulcer}, \textsc{hush}, \textsc{oak}, \textsc{lion}, \textsc{obtain}, \textsc{insult}, \textsc{sixth}, \textsc{azariah}, \textsc{galileans}, \textsc{inhabited}, \textsc{overwhelming}, \textsc{jerimoth}, \textsc{tabernacles}, \textsc{tophet}

Other concept lists are made available online~\url{https://anonymous.4open.science/r/PersistenceAndDiffusibility-018F/data/wordlists/}.

\begin{figure}[ht]
    \centering
    \includegraphics[width=0.5\textwidth]{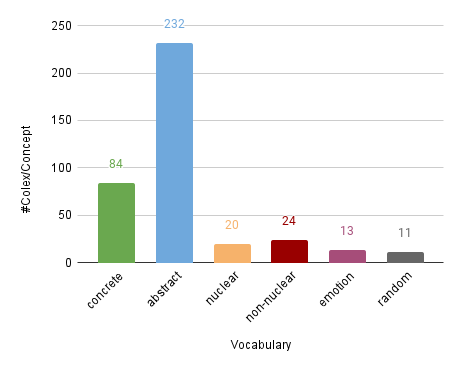}
    \caption{Average number of colexification patterns per each concept in ColexNet+.}
    \label{fig:dist_colex}
\end{figure}

\begin{table}[ht]
    \centering
    \resizebox{0.4\textwidth}{!}{
    \begin{tabular}{cccc}
    \hline 
         Vocabulary & \#SEED& \#Concept & \#Colex.Patterns  \\\hline 
        \textsc{nuclear}& 40 & 3051	 & 30887 \\
        \textsc{non-nuclear} & 60 & 3206 & 37969 \\
        \textsc{emotion} & 23 & 2807 & 	18677 \\
        \textsc{random} & 60 & 2797	 & 15838 \\\hline
        \textsc{concrete} & 1094 & 1024 & 	43127 \\
        \textsc{abstract}  & 1814 & 1781 & 	206406 \\\hline 
     \end{tabular}}
    \caption{The distribution of concepts and colexification patterns for vocabularies from ColexNet+ across 1334 languages. For nuclear, non-nulcear, emotion, and random colexifications, only either of the pair of concepts involved in each colexification pattern has to be from the regarding vocabulary. For concrete and abstract colexification, both concepts involved in a colexification pattern have to be concrete or abstract, and there is a threshold ($\geq 3$) to control rare colexification patterns across languages. }
    \label{tab:dist_concept_colex }
\end{table}

\begin{table*}[ht]
    \centering
    \begin{tabular}{ccccc}
    \hline 
         Vocabulary &   \#Concept & \#Colex.Patterns & \#Lang & \#Colex/Lang  \\\hline 
        \textsc{nuclear} & 2042 &	8957 &	1322	 & 6493 \\
        \textsc{non-nuclear}  & 2158	&10715	&1322	&7387\\
        \textsc{emotion} & 1757	& 6060 &	1322 & 	5782\\
        \textsc{random}  &1826	&4007	&1327	&4769\\\hline
        \textsc{concrete}   &798 &	12132 &		1327	&	5098\\
        \textsc{abstract}   & 1505 & 	65771 & 	1327	& 4604\\\hline 
         \textsc{aff.abstract} & 114 &	863 &	1325 &	6591 \\
        \textsc{aff.concrete} &	41	& 68	& 1325	& 3760 \\\hline
  
     \end{tabular}
    \caption{The distribution of concepts and colexification patterns for vocabularies in Bible data. }
    \label{tab:dist_colex_bible }
\end{table*}

\begin{table}[ht]
    \centering
    \resizebox{0.5\textwidth}{!}{
    \begin{tabular}{ccccc}
    \hline 
         Vocabulary &   \#Concept & \#Colex.Patterns & \#Lang & \#Colex/Lang  \\\hline 
        \textsc{nuclear} & 1779  &	5000	 & 1322 &	10931 \\
        \textsc{non-nuclear}  &	1771 &		5000 &		1322	& 14341\\
        \textsc{emotion} & 1670	 & 5439	& 1322	& 5037\\
        \textsc{random}  &1742	  & 3772 &	1327	 & 4406\\\hline
        \textsc{concrete}   &703	& 8630	& 1326 &	4604\\
        \textsc{abstract}   & 1110	& 8630 &	1327 &	23592\\\hline 
        \textsc{aff.abstract} & 58 &	68 &	1316 &	65597 \\
        \textsc{aff.concrete} &	41	& 68	& 1182	& 3760 \\\hline

     \end{tabular}
     }
    \caption{The distribution of concepts and colexification patterns sampled for testing hypotheses for differential persistence and diffusibility. }
    \label{tab:dist_colex_bible_another }
\end{table}

\appendixsection{Language Contact Graphs}\label{sec:languge_contact_appendix}


\begin{figure}[tb]
    \centering
    \includegraphics[width=0.5\textwidth]{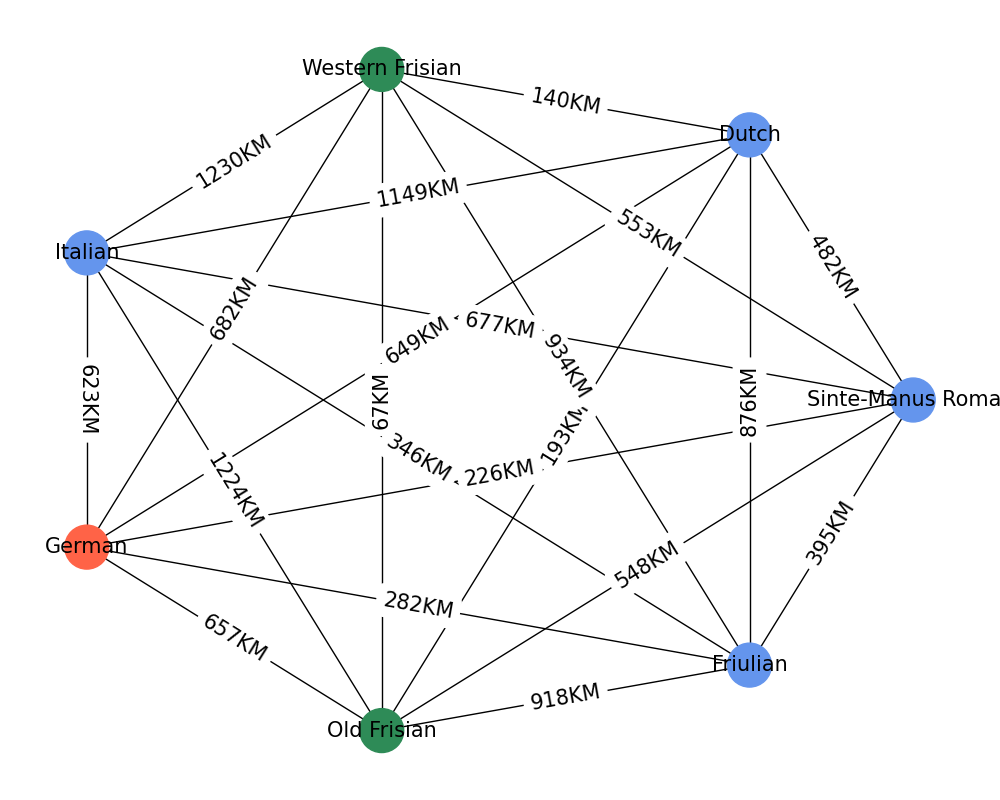}
    \includegraphics[width=0.5\textwidth]{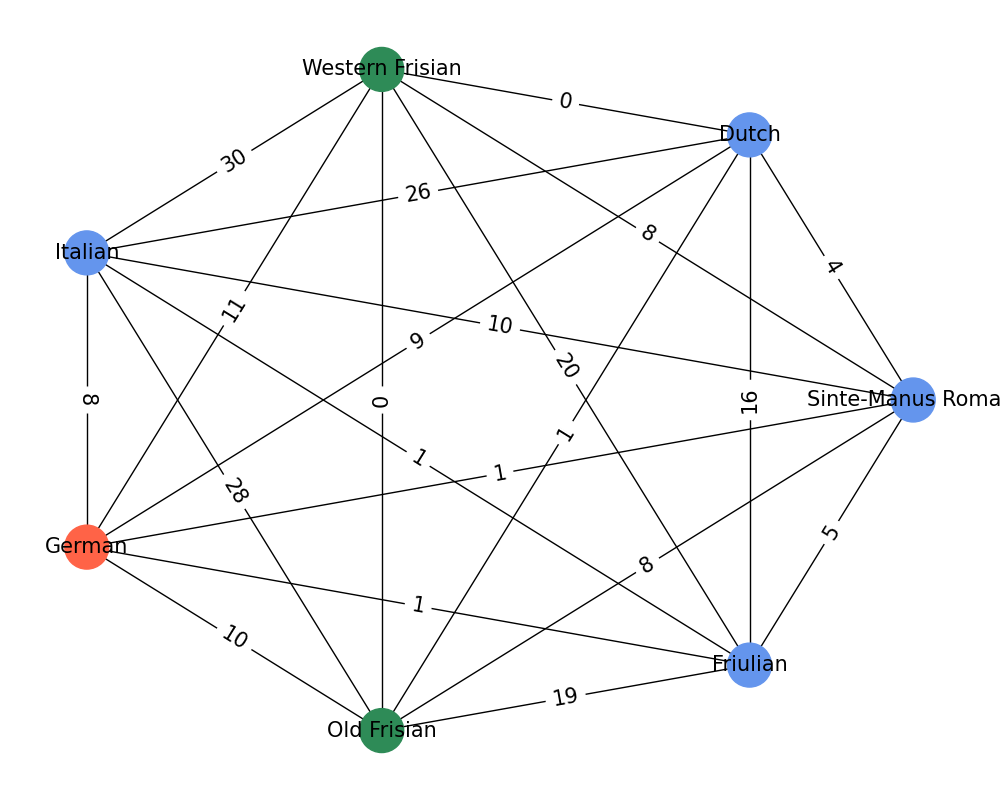}
    \caption{Language contact graph of a sub-graph of German. The colors represent neighbourhood between languages: center language (red), non-neighouring languages (green) and neighbouring languages (blue). (Top) The numbers on the edges represent the geographical distances between each pair of languages. (Bottom) The numbers on the edges represent the contact languages in between each pair of languages }
    \label{fig:geodist_graph}
\end{figure}

In their research,~\citet{GastKoptjevskajaTamm+2022+403+438} create a contact network using geographical data of languages. They visualize the positions of major languages using points and polygons, while minor languages are represented as points only. The researchers classify languages as "neighbours" if the distance between their polygons is within 50km for major languages and 100km for both minor and major languages. Previous researches have not provide a standard diagnostic measure for calculating contact languages and neighbouring languages. However, their research focuses solely on European countries and lacks a standardized diagnostic measure for calculating contact and neighbouring languages. To address the large number of languages, we adopt the measurement proposed by~\citep{dryer2018order} (cf. Section~\ref{language_graph}). Fig.~\ref{fig:geodist_graph} shows the contact graphs based on geographical distances, number of contact languages, and neighbours, with German as the central language.

\begin{figure*}[tb]
    \centering
    \includegraphics[width=0.4\textwidth]{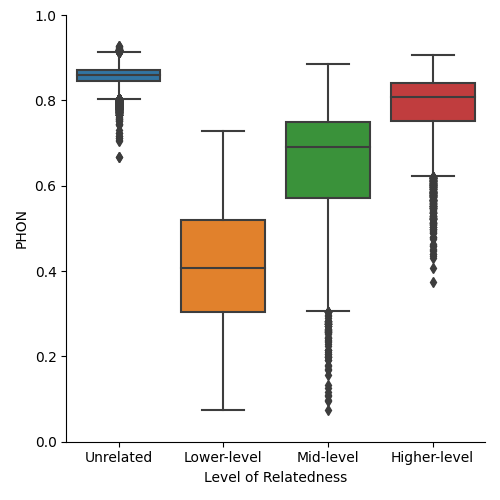}
    \includegraphics[width=0.4\textwidth]{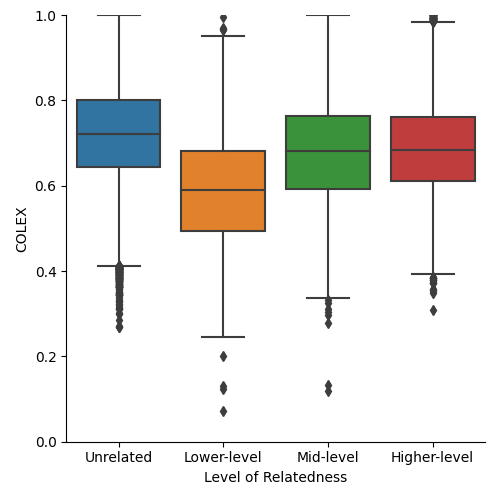}
    \caption{Data Distribution of (left) Phonological Distance and (right) Colexificaiton Distance based on \textsc{Nuclear} vocabulary, by the level of genealogical relatedness.}
    \label{fig:data_dist_plot}
\end{figure*}

\appendixsection{Data Distribution of \textsc{phon} and \textsc{colex}}\label{data_dist}
The data distribution of phonological distance and colexification distance of each level of genealogical relatedness based on \textsc{nuclear} vocabulary is shown in Fig.~\ref{fig:data_dist_plot}. The phonological distance is consistently higher in higher-level related and unrelated languages while lower in lower-level related languages (Fig.~\ref{fig:data_dist_plot} left). The colexification distance is relatively evenly distributed across the levels of genealogical relatedness, except being lower in lower-level related languages (Fig.~\ref{fig:data_dist_plot} right).

\appendixsection{Coefficients Among Variables}\label{sec:coeff_colex_phon}

\begin{figure*}[ht]
    \centering
    \includegraphics[width=0.8\textwidth]{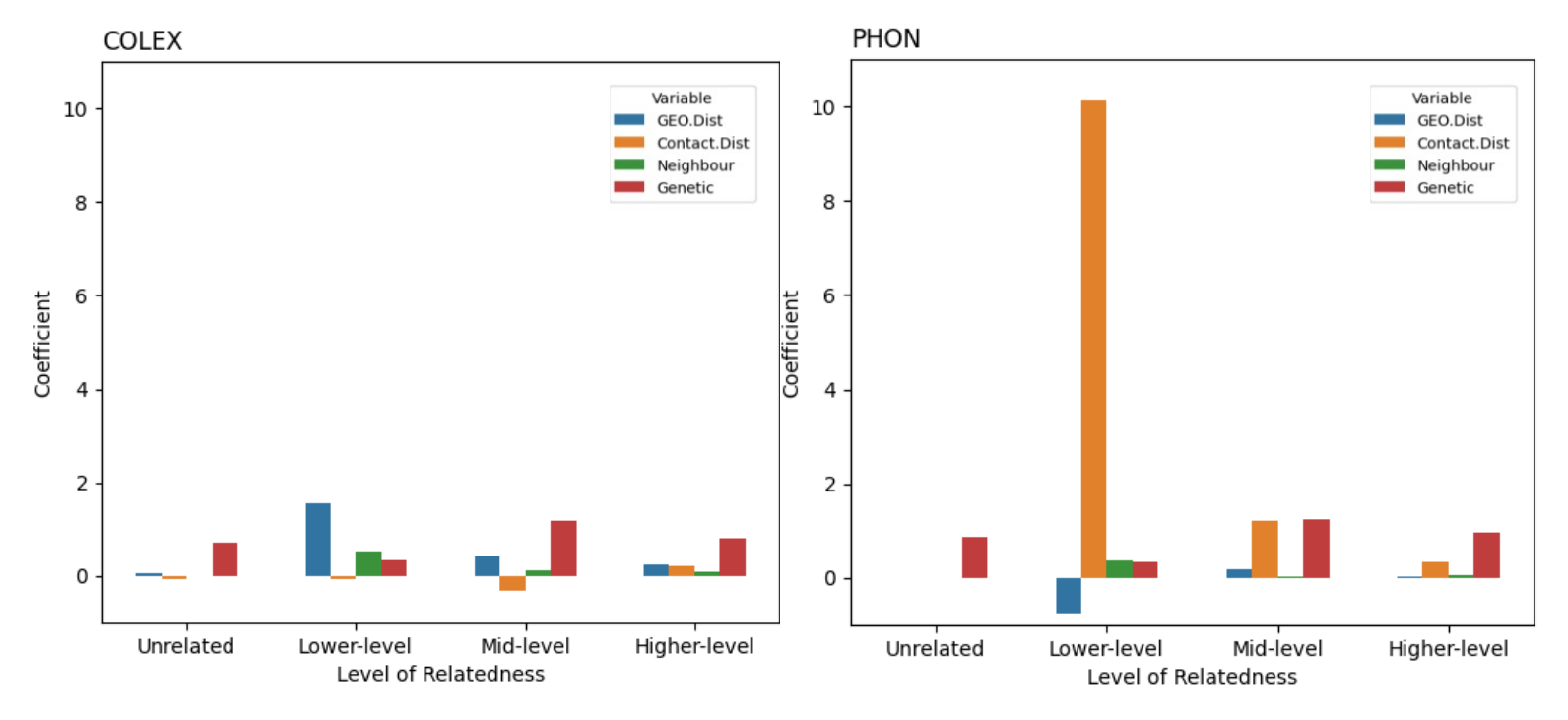}
    \caption{Coefficients among variables for \textsc{phon} and \textsc{colex}. The Confidence interval at 95\% are minimal, and for all $p<0.001$. The values that are invisible in the plots are near to 0.}
    \label{fig:phon_colex_ols}
\end{figure*}

To overview the effect size of variables as indicators for both colexification distances and phonological distances, we use ordinary least squares to calculate coefficients between them and variables. For colexification distances, there are 1,313 languages and 861,328 languages; for phonological distance, there are 1,558 languages and 722,294 language pairs. As shown in Fig.~\ref{fig:phon_colex_ols} (left), for unrelated, lower-level, and mid-level related languages, there is a negative correlation between language contact distance and colexification patterns. While Fig.~\ref{fig:phon_colex_ols} (right) shows that, the correlation between language contact distance and phonological distance is more significant for closely related (lower-level) languages than for more remotely related languages and especially unrelated languages. 

\appendixsection{Pairwise Language Family ARIs on Colexification Patterns }\label{pairwise_aris}

\begin{figure}[ht]
    \centering
    \includegraphics[width=0.45\textwidth]{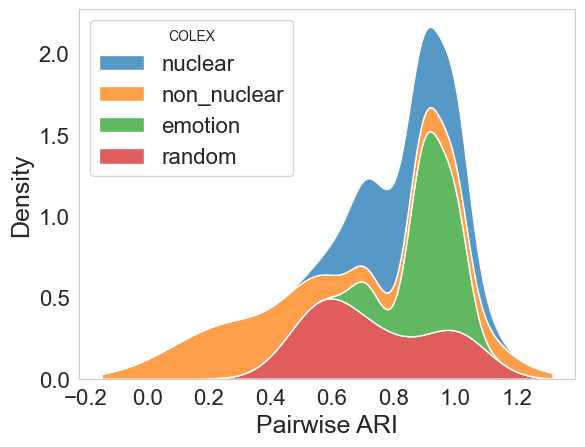}
    \caption{The distribution of pairwise ARI values for semantic distances involving \textsc{nuclear}, \textsc{non-nuclear}, \textsc{emotion} and \textsc{random} across top 5 language families with the most languages, i.e., Atlantic-Congo, Austronesian, Nuclear Trans New Guinea, Indo-European, and Afro-Asiatic.}
    \label{fig:aris}
\end{figure}

ARIs are used to quantify the agreement in comunity structure, which indicate the similarity of two network's community. Negative ARIs indicate that two netowrk's community partitions vary more than would be expected by chance, ARI values of 0 indicate that the variation comes by chance, while ARI values naer 1 reflect high agreement in community structure between two networks. To test the whether variations in colexification patterns involving different concepts merely vary by chance, we calculate pairwise ARIs across top 5 language families with the most number of languages, i.e., Atlantic-Congo, Austronesian, Nuclear Trans New Guinea, Indo-European, and Afro-Asiatic. As shown in Fig.~\ref{fig:aris}, for the colexifiation patterns involving \textsc{nuclear}, \textsc{non-nuclear}, \textsc{emotion} and \textsc{random}, the ARIs are all higher than 0.5, which could mean that the colexification patterns involving all these concept sets are permissive in the Bible data across language families. While \textsc{random} has much lower ARI values than the others, indicating more semantic variability. There is a differential semantic variability from \textsc{nuclear}, \textsc{non-nuclear} to \textsc{emotion}, from the lower to the higher, considering the different ARI values, which also partially corroborates the findings from \citet{jackson-2019}, where emotion concepts show much higher semantic variability than color concepts (part of the core concepts).

\section{Semantic Distances vs. Language Contact Distances for lower-related Languages}\label{sec:colex_branches}

\begin{figure}[ht]
    \centering
        \includegraphics[width=0.3\textwidth]{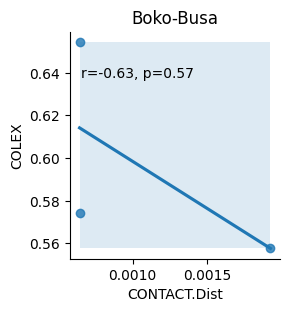}
    \includegraphics[width=0.3\textwidth]{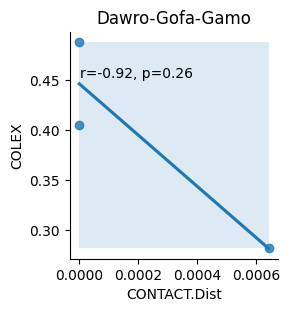}
    \includegraphics[width=0.3\textwidth]{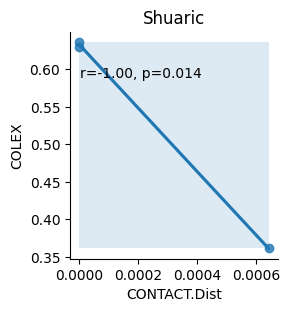}
    \includegraphics[width=0.3\textwidth]{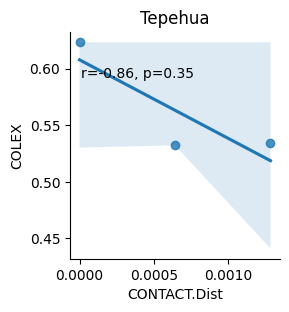}
    \caption{
    Examples of semantic distances plotted against language contact distances for language pairs related at lower-level, i.e., in the same branch, where they are negatively correlated. }
    \label{fig:colex_contact_lm_negative}
    \end{figure}

\begin{figure}[ht]
    \centering
     \includegraphics[width=0.3\textwidth]{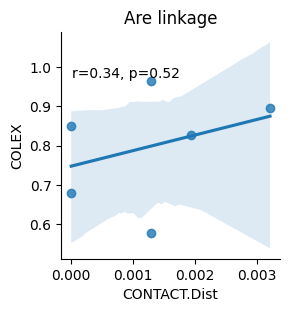}
         \includegraphics[width=0.3\textwidth]{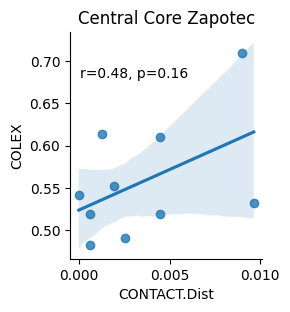}

    \includegraphics[width=0.3\textwidth]{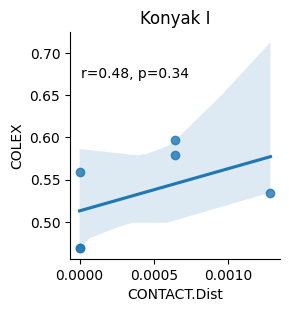}
    \includegraphics[width=0.3\textwidth]{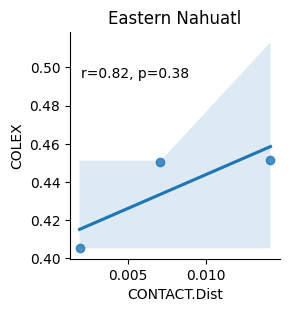}
    \caption{Examples of semantic distances plotted against language contact distances for language pairs related at lower-level, i.e., in the same branch, where they are positively correlated. }
    \label{fig:colex_contact_lm_positive}
    \end{figure}

The relationship between semantic distances and language contact distances of languages on the same branch is shown in Fig.~\ref{fig:colex_contact_lm_negative} and Fig.~\ref{fig:colex_contact_lm_positive} (cf. Section~\ref{subsec:phon_colex_geo}). See all the plots in the GitHub repository~\url{https://anonymous.4open.science/r/PersistenceAndDiffusibility-018F/plots/colex_geocontact/}.

\end{document}